\documentclass[preprint,12pt]{elsarticle}

\usepackage{float}
\usepackage{natbib}
\usepackage{graphicx}
\usepackage{subfigure}
\usepackage{epsfig}
\usepackage{multirow}
\usepackage{multicol}
\usepackage{makecell}
\usepackage{array}
\usepackage{diagbox}
\usepackage{amssymb}
\usepackage{color}
\usepackage{algorithm} 
\usepackage{algpseudocode}

\usepackage{mathrsfs}
\usepackage{amsmath}
\usepackage{amssymb}
\usepackage{extarrows}
\usepackage{setspace}
\usepackage{hyperref}

\usepackage{epstopdf}  
\usepackage{amsthm}
\usepackage{epsfig}
\usepackage{dsfont}

\journal{Neurocomputing}

\begin{document}

\begin{frontmatter}

\title{Exponential Discriminative Metric Embedding\\
in Deep Learning}


\author[a]{Bowen Wu \corref{cor}}
\author[b]{Zhangling Chen}
\author[c]{Jun Wang}
\author[b]{Huaming Wu}
\address[a]{Center for Combinatorics, Nankai University, Tianjin 300071, China}
\address[b]{Center for Applied Mathematics, Tianjin University, Tianjin 300072, China}
\address[c]{School of Mathematics, Tianjin University, Tianjin 300072, China}
\cortext[cor]{Corresponding author. \\ \indent \indent E-mail address: wbw@mail.nankai.edu.cn (B. Wu).}

\begin{abstract}

With the remarkable success achieved by the Convolutional Neural Networks (CNNs) in object recognition recently, deep learning is being widely used in the computer vision community. Deep Metric Learning (DML), integrating deep learning with conventional metric learning, has set new records in many fields, especially in classification task. In this paper, we propose a replicable DML method, called Include and Exclude (IE) loss, to force the distance between a sample and its designated class center away from the mean distance of this sample to other class centers with a large margin in the exponential feature projection space. With the supervision of IE loss, we can train CNNs to enhance the intra-class compactness and inter-class separability, leading to great improvements on several public datasets ranging from object recognition to face verification. We conduct a comparative study of our algorithm with several typical DML methods on three kinds of networks with different capacity. Extensive experiments on three object recognition datasets and two face recognition datasets demonstrate that IE loss is always superior to other mainstream DML methods and approach the state-of-the-art results.

\end{abstract}


\end{frontmatter}


\section{Introduction}

Recently, Convolutional Neural Networks (CNNs) are continuously setting new records in classification aspect, such as object recognition \cite{krizhevsky2012imagenet,simonyan2014very,szegedy2015going,he2015deep}, scene recognition \cite{zhou2014object,zhou2014learning}, face recognition \cite{taigman2014deepface,lu2014surpassing,sun2014deep,sun2015deeply,schroff2015facenet,wen2016discriminative}, age estimation \cite{levi2015age,liu2016group} and so on. Facing the more and more complex data, the deeper and wider CNNs tend to obtain better accuracies. Meanwhile, many troubles will show up, such as gradient saturating, model overfitting, parameter augmentation, etc. To solve the first problem, some non-linear activations \cite{goodfellow2013maxout,he2015delving,shang2016understanding} have been proposed. Considerable efforts have been made to reduce model overfitting, such as data augmentation \cite{krizhevsky2012imagenet,cirecsan2011high}, dropout \cite{hinton2012improving,krizhevsky2012imagenet}, regularization \cite{goodfellow2013maxout,wan2013regularization}. Besides, some model compressing methods \cite{han2015deep,sun2016sparsifying} have largely reduced the computing complexity of original models, with the performance improved simultaneously.

In general object recognition, scene recognition and age estimation, the identities of the possible testing samples are within the training set. So the training and testing sets have the same object classes but not the same images. In this case, softmax classifier is often used to designate a label to the input.

\begin{figure}[htp!]\centering
 \includegraphics[height=6.5cm,width=13.5cm]
 {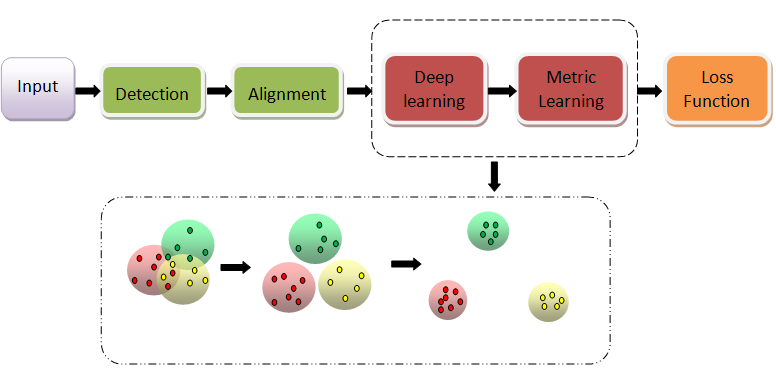}
 \caption{The typical framework of face recognition. The process of deep feature learning and metric learning is shown in the second row.}\label{fig1}
\end{figure}

For face recognition, the deeply learned features need to be not only separable but also discriminative. It can be roughly divided into two aspects, namely face identification and face verification. The former is the same as object recognition, the training and testing sets have the same face identities, aims at classifying an input image into a large number of identity classes.  Face verification is to classify a pair of images as belonging to the same identity or not (i.e. binary classification). Since it is impractical to pre-collect enough number of all the possible testing identities for training, face verification is becoming the mainstream in this field. As clarified by DeepID series \cite{sun2014deep,sun2014deep1,sun2015deeply}: classifying all the identities simultaneously instead of binary classifiers for training can make the learned features more discriminative between different classes. So we decide to use the joint supervision of softmax classifier and metric loss function to train and the verification signal of feature similarity discriminant to test as shown in Section 4.3. Fig.~\ref{fig1} illustrates the general face recognition pipeline, which maps the input images to the discriminative deep features progressively, then to the predicted labels.

A recent trend towards deep learning with more discriminative features is to reinforce CNNs with better metric loss functions, namely Deep Metric Learning (DML), such that the intra-class compactness and inter-class separability are simultaneously maximized. Inspired by this idea, many metric learning methods have been proposed. It can be traced back to early subspace face recognition methods such as Linear Discriminant Analysis (LDA) \cite{belhumeur1997eigenfaces}, Bayesian face \cite{moghaddam2000bayesian}, and unified subspace \cite{wang2004unified}. For example, LDA aims at maximizing the ratio between inter-class and intra-class variations by finding the optimal projection direction. Some metric learning methods \cite{chen2012bayesian,weinberger2005distance,kan2013adaptive} have been proposed to project the original feature space into another metric space, such that the features of the same identity are close and those of different identities stay apart. Subsequent contrastive loss \cite{sun2014deep1} and triplet loss \cite{schroff2015facenet} have witnessed their success in face recognition.

Interestingly, closely related to DML is the Learning to Hash, which is one of the major solutions to nearest neighbor search problem. Given the high dimensionality and high complexity of multimedia data, the cost of finding the exact nearest neighbor is prohibitively high. Learning to Hash, a data-dependent hashing approach, aims to learn hash functions from a specific dataset so that the nearest neighbor search result in the hash coding space is as close as possible to the search result in the original space, significantly improving the search efficiency and space cost. The main methodology of Learning to Hash is similarity preserving, i.e., minimizing the gap between the similarities computed in the original space and the similarities in the hash coding space in various forms. \cite{song2015supervised} utilizes linear LDA with trace ratio criterion to learn hash functions, where the pseudo labels and the hash codes are jointly learned. \cite{gao2015scalable} proposes a semi-supervised deep learning hashing method for fast multimedia retrieval, to simultaneously learn a good multimedia representation and hash function. More comprehensive survey about dimension reduction and using different similarity preserving algorithms to hashing can be found in \cite{gao2017learning,wang2017survey}. Surprisingly, most of the similarity metric loss functions could be used for Learning to Hash.

Because of the large scale of training set, it is unreasonable to address all of them in each iteration. Mini-batch based Stochastic Gradient Descent (SGD) algorithm \cite{lecun1998gradient} doesn't reflect the real distribution of the total training set, so a superior sampling strategy becomes very important to the training process. Besides, selecting appropriate pairs or triplets like previous may dramatically increase the number of training samples. As a result, it is inevitably hard to converge to an optimum steadily. In this paper, we propose a novel well-generalized metric loss function, named Include and Exclude (IE) loss, to make the deeply learned features more discriminative between different classes and closer to each other between images of the same class. This idea is verified by Fig.~\ref{fig2} in Section 3.1. Obviously, the inter-class distance is away from the intra-class distance with a large margin. When training, we learn a center for each class like center loss \cite{wen2016discriminative} does. Subsequently, we show that center loss is a variant of the special case of our method. There is another parameter $\sigma^{2}$ to regularize the distance between the features and their corresponding class centers. Furthermore, we use a hyperparameter $Q$ to control the number of valuable inter-class distances to accelerate the convergence of our model. We simultaneously use the supervision signals of softmax loss and IE loss to train the network. Extensive experiments on object recognition and face verification validate the effectiveness of IE loss. Our method significantly improves the performance compared to the original softmax method, and competitive with other nowadays mainstream DML algorithms. The main contributions are summarized as follows:
\begin{itemize}
  \item  To the best of our knowledge, we are the first to practice the idea of enforcing the mean inter-class distance larger than the intra-class distance with a margin in the exponential feature projection space, as opposed to the distance between a sample and its nearest cluster centers in magnet loss \cite{rippel2016metric}, avoiding the large intra-class distances.
  \item  Instead of some off-line complicated sampling strategies, our DML method can achieve a satisfactory result only using the mini-batch based SGD, greatly simplifying the training process.
  \item  To achieve a better performance rapidly, we introduce a hyperparameter $Q$ to restrict the number of nearest inter-class distances in each mini-batch to accelerate the convergence of our model.
  \item We do extensive experiments on several common datasets, including MNIST, CIFAR10, CIFAR100, Labeled Faces in the Wild (LFW) and YouTube Faces (YTF), to verify the effectiveness, robustness and generalization of IE loss.
\end{itemize}

\section{Related work}

In recent years, deep learning has been successfully applied in computer vision and other AI domains, such as object recognition \cite{szegedy2015going}, face recognition \cite{schroff2015facenet}, image retrieval \cite{Song2018binary, song2018quantization}, speech recognition \cite{graves2013speech} and natural language processing \cite{socher2013recursive}. Most of the time, deep learning models are prone to be deeper and wider. But more complicated deep networks are accompanied by larger training set, model overfitting and costly computational overhead. Considering these, there produce some new DML methods, which concatenate the conventional metric learning losses to the end of the deeply learned features. In classification aspect, DML generally aims at mapping the originally learned features into a more discriminative feature space by maximizing the inter-class variations and minimizing the intra-class variations. To some degree, a properly chosen metric loss function would make the training easy to converge to an optimal model without too much training data. We will briefly discuss some typical DML methods below.

Sun et al. \cite{sun2014deep1} encourage all faces of one identity to be projected onto a single point in the embedding space. They use an ensemble of 25 networks on different face patches to get the final concatenated features. Both PCA and Joint Bayesian classifier \cite{chen2012bayesian} are used to achieve the final performance of $99.47\%$ on LFW. The loss function is mainly based on the idea of contrastive loss, which minimizes the intra-class distance and enforces the inter-class distance larger than a fixed margin.

Schroff et al. \cite{schroff2015facenet} employ the triplet loss, which stems from LMNN \cite{weinberger2005distance}, to encourage a distance constraint similar to the contrastive loss. Differently, the triplet loss requires a triple of training samples as input at a time, not a pair. The triplet loss minimizes the distance between an anchor sample and a positive sample, and maximizes the distance between the anchor sample and a negative sample, in order to make the inter-class distance larger than the intra-class distance by a margin relatively. They also use the so far largest training database about 200M face images, and set an insurmountable record on LFW of $99.63\%$.

Rippel et al. \cite{rippel2016metric} propose a novel magnet loss, which is explicitly designed to maintain the distribution of different classes in feature space. In terms of computational performance, it alleviates the training inefficiency of the traditional triplet loss, which is verified from classification task to attribute concentration. But, the complicated off-line sampling strategy makes it too difficult to reproduce. In addition, the intra-class distribution maintaining by local clusters would impair the inter-class separability in general classification tasks, especially in face recognition.

\section{The proposed approaches}
We first clarify the notations which will be used in subsequential sections. Let us assume the training set consists of $M$ input-label pairs $\mathcal{D}=\{x_{n},y_{n}\}^{M}_{n=1}$ belonging to $C$ classes. We consider a parameterized map $f(x_{n},\Theta), n=1,\cdots, M$, and $\Theta$ are the model parameters. In this work, the transformation is selected as some complex CNN architectures. We further define $C(f_{n})$ as the class label of feature $f_{n}$, and $\mu_{C(f_{n})}$ as the corresponding class center.

\subsection{Some existing methods}

In this section, some existing superior DML methods are first presented.

$\mathbf{Triplet\   Loss}\  \   $ Schroff et al. \cite{schroff2015facenet} have verified the effectiveness of triplet loss with a large training set. But the exponentially increased computational complexity of training examples and the difficulty of convergence impede its general application. The formula is as follows:
\begin{equation}
\centering
\mathcal{L}(\Theta)=\sum_{i=1}^{M}\left\{\|f(x^{a}_{i})-f(x^{p}_{i})\|^{2}_{2}-\|f(x^{a}_{i})-f(x^{n}_{i})\|^{2}_{2}+\alpha\right\}_{+}.
\end{equation}
Here, $x^{a}_{i}$, $x^{p}_{i}$ and $x^{n}_{i}$ refer to the anchor, positive and negative images in a triplet, respectively. $\alpha$ is the predefined margin.

$\mathbf{L}$-$\mathbf{Softmax\ Loss}\  \   $Liu et al. \cite{liu2016large} achieve a flexible learning objective with adjustable difficulty, by altering the classification angle margin between classes. Although the relatively rigorous learning objective with adjustable angle margin can avoid overfitting, the difficult convergence hinders its generalization to many other deep networks. It is crucial to continuously adjust the component weight between softmax and L-Softmax to guarantee the progressing of training.
  \begin{equation}
  \centering
\mathcal{L}(\Theta)=-\frac{1}{M}\sum^{M}_{i=1}\log\left(\frac{exp({\|W_{y_{i}}\|\|x_{i}\|\psi(\theta_{y_{i}})})}{exp({\|W_{y_{i}}\|\|x_{i}\|\psi(\theta_{y_{i}})})+\sum_{j\neq y_{i}}exp({\|W_{j}\|\|x_{i}\|cos(\theta_{j})})}\right).
  \end{equation}
It generally requires that
  \begin{equation}
  \centering
  \psi(\theta)=
  \begin{cases}
   \cos(m\theta),  &   0\leq\theta\leq\frac{\pi}{m}\\
   \mathcal{D}(\theta), &  \frac{\pi}{m}<\theta\leq\pi
  \end{cases}
  \end{equation}
where $W$ is the weight matrix of the fully connected layer before softmax layer, and $W_{y_{i}}$ is the $y_{i}$-th column of $W$. $\theta_{y_{i}}$ is the angle between $x_{i}$ and its corresponding weight vector $W_{y_{i}}$, and $m$ is an integer to control the learning objective. Meanwhile, $\mathcal{D}(\theta)$ must be monotonically decreased to satisfy the requirement for any $\theta$.

$\mathbf{Center\   Loss}\  \   $ Wen et al. \cite{wen2016discriminative} propose a new loss function, which regards the distance of a sample away from its corresponding class center as the objective penalization. The joint supervision of center loss and softmax loss makes this approach outperform most existing best results on some face recognition benchmark databases.
  \begin{equation}
  \centering
\mathcal{L}(\Theta)=\frac{1}{2M}\sum^{M}_{i=1}\|f(x_{i})-\mu(f(x_{i}))\|^{2}_{2},
  \end{equation}
where $\mu(f(x_{i}))$ is the class center of $f(x_{i})$.

\subsection{IE Loss}

As clarified in \cite{rippel2016metric}, magnet loss liberates us from the unreasonable prior target neighbourhood assignments, and divides each class into several clusters, aims at maintaining the distributions of different classes in the representation space. As a result, the similar samples in different classes may be closer than that in the same classes. Specifically, intra-class variations may be larger than inter-class variations in object recognition and face recognition. Thus some local distribution maintaining loss functions like magnet loss will not bring so many benefits to the practical classification tasks. Despite the great performance on LFW by triplet loss on GoogLeNet \cite{szegedy2015going}, its training ineffectiveness and the exponentially increased training samples hinder the widespread application to generic classification tasks.

Considering the difficulty of magnet loss to reproduce and the disadvantages mentioned above, we propose a replicable DML method, called IE loss, to learn the discriminative features. We calculate all the distances between a sample and other class centers in a mini-batch to take of advantage of batch information, as compared to the pair/triplet samples like previous. The objective is initially defined as follows:

\begin{equation}
\centering
\mathcal{L}(\Theta)=\frac{1}{M}\sum^{M}_{n=1}\left\{-log\frac{exp({-\frac{1}{2\sigma^{2}}\|f_{n}-\mu_{C(f_{n})}\|^{2}_{2}-\alpha})}{\sum_{c\neq C(f_{n})}exp({-\frac{1}{2\sigma^{2}}\|f_{n}-\mu_{c}\|^{2}_{2}})}\right\}_{+},
\end{equation}
where $\{\cdot\}_{+}$ is the hinge loss function, $\alpha$ is a predefined margin hyperparameter, $\sigma^{2}=\frac{1}{M-1}\sum_{n\in \mathcal{D}}\|f_{n}-\mu_{C(f_{n})}\|^{2}_{2}$ is the variance of examples away from their respective class centers in the feature space. When training, the class center $\mu_{C(f_{n})}$ and variance $\sigma^{2}$ should update together with the deep feature $f_{n}$. This means we should use the entire training set in each iteration. Obviously, it is impractical. So we decide to employ the mini-batch based SGD algorithm to update the parameters. The denominator in log part is computed by summing all the inter-class distances between a sample and other class centers appear in the mini-batch. This approach seems to be a natural choice with the probability interpretation, the same to softmax loss.

Some existing similar DML methods express that a sample quite far away from the corresponding class center should vanish from its term in our objective, approximating the denominator of Equation 5 with a small number of nearest classes. Variance standardization also renders the objective invariant to the characteristic length scale of the problem. Whereas, all these benefits are based on a superb neighborhood sampling strategy for each class to keep the local distribution. Different from the strategy exploited in \cite{rippel2016metric} which sampling the nearest $K$ clusters in each class, we decide to use the $Q$ nearest class centers to obtain the objective. The improved objective loss function is formulated as follows:

\begin{equation}
\centering
\mathcal{L}(\Theta)=\frac{1}{M}\sum^{M}_{n=1}\left\{-log\frac{exp({-\frac{1}{2\sigma^{2}}\|f_{n}-\mu_{C(f_{n})}\|^{2}_{2}-\alpha})}{\sum_{c=1,c\neq C(f_{n})}^{Q}exp({-\frac{1}{2\sigma^{2}Q}\|f_{n}-\mu_{c}\|^{2}_{2}})}\right\}_{+},
\end{equation}
where $Q$ is an effectively selected number of different inter-class distances between a sample and other class centers in a mini-batch, and these distances are sorted in ascending order. We can choose a proper $Q$ according to different training datasets to acquire the best performance. One can notice that the sophisticated off-line nearest clusters sampling strategy is avoided, and the mini-batch based SGD works well for our training. Besides, the too large inter-class distances are removed to accelerate the convergence, which is especially valid for the datasets with many classes. Subsequent results will show that the proposed method can greatly improve the training efficiency without sacrificing speed, since these auxiliary loss layers are removed in the classification step.

When we set $Q=1$ and $\sigma^{2}=0.5$, Eq.(6) immediately reduces to Eq.(7).

\begin{equation}
\centering
\mathcal{L}(\Theta)=\frac{1}{M}\sum^{M}_{n=1}\left\{\|f_{n}-\mu_{C(f_{n})}\|^{2}_{2}+\alpha-\min_{c\neq C(f_{n})}\|f_{n}-\mu_{c}\|^{2}_{2}\right\}_{+}.
\end{equation}
It is clear that this formula is a variant of the efficient center loss and triplet loss. This loss function seems more appropriate to reflect the characteristics of our proposed method. It apparently forces the minimum inter-class distance larger than the intra-class distance with a margin $\alpha$.

\begin{figure}[h]
\centering
 \includegraphics[height=7cm,width=14cm]
 {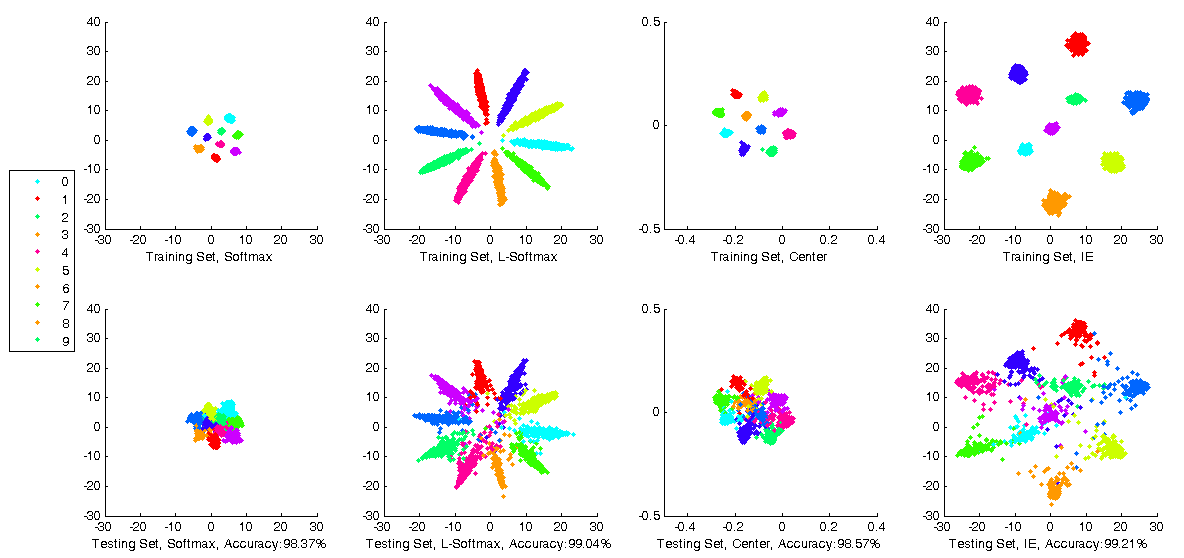}
 \caption{Visualization of the deeply learned 2D features on training and testing sets of MNIST, regarding softmax loss, L-Softmax loss, center loss and IE loss, respectively. The points with different colors correspond to the features from different classes.}\label{fig2}
\end{figure}

The effectiveness of our method is shown in Fig.\ref{fig2}. The visualization of 2-D features on training and testing sets sufficiently reflects the relative intra-class compactness and inter-class separability of IE loss, compared to softmax loss. One can also find that L-Softmax loss obviously amplifies the angle of features between different classes, and center loss seriously shrinks the intra-class distances such that the deeply learned features are discriminative in a small subspace.

Considering the classical back-propagation algorithm, the entire parameter updating process of IE loss is summarized in Algorithm \ref{algorithm1}. Softmax loss is incorporated to accelerate the converge of our training process. $\lambda$ is the weighting parameter between softmax loss and IE loss in our final objective, to keep the balance between these two supervision symbols.

\begin{algorithm}[H]
  \caption{The parameter updating algorithm of IE loss.}
  \label{algorithm1}
  \begin{algorithmic}[1]
    \Require
      training set $\mathcal{D}=\{x_{n}, y_{n}\}^{M}_{n=1}$, initialized parameters $\theta_{c}$ in convolutional layers, $W$, $\sigma^{2}$ and $\mu_{q}(q=0,1,\ldots,Q)$ in loss layer where $q=0$ corresponds to the case of $\mu_{C(f_{n})}$, hyperparameters $\alpha$ and $\lambda$, learning rate $\eta^{t}$ and total iterative steps $T$.
    \Ensure
      model parameters $\theta_{c}$.
    \For{$t=1,2,\cdots,T$}
    \State compute the loss function
    \State $\mathcal{L}^{t}=\mathcal{L}^{t}_{softmax}+\lambda \mathcal{L}^{t}_{IE}$
    \State compute the gradients
    \State $\frac{\partial\mathcal{L}^{t}}{\partial f_{n}^{t}}=\frac{\partial\mathcal{L}^{t}_{softmax}}{\partial f_{n}^{t}}+\lambda\frac{\partial\mathcal{L}^{t}_{IE}}{\partial f_{n}^{t}}$
    \State $\frac{\partial\mathcal{L}^{t}}{\partial W^{t}}=\frac{\partial\mathcal{L}^{t}_{softmax}}{\partial W^{t}}+\lambda\frac{\partial\mathcal{L}^{t}_{IE}}{\partial W^{t}}=\lambda\frac{\partial\mathcal{L}^{t}_{IE}}{\partial W^{t}}$
    \State $\frac{\partial\mathcal{L}^{t}}{\partial \mu_{q}^{t}}=\frac{\partial\mathcal{L}^{t}_{softmax}}{\partial \mu_{q}^{t}}+\lambda\frac{\partial\mathcal{L}^{t}_{IE}}{\partial \mu_{q}^{t}}=\lambda\frac{\partial\mathcal{L}^{t}_{IE}}{\partial \mu_{q}^{t}}$
    \State $\frac{\partial\mathcal{L}^{t}}{\partial \sigma^{2}_{t}}=\frac{\partial\mathcal{L}^{t}_{softmax}}{\partial \sigma^{2}_{t}}+\lambda\frac{\partial\mathcal{L}^{t}_{IE}}{\partial \sigma^{2}_{t}}=\lambda\frac{\partial\mathcal{L}^{t}_{IE}}{\partial \sigma^{2}_{t}}$
    \State update parameters
    \State $W^{t+1}=W^{t}-\eta^{t}\cdot\large{\frac{\partial\mathcal{L}^{t}}{\partial W^{t}}}=W^{t}-\eta^{t}\cdot\lambda\cdot\large{\frac{\partial\mathcal{L}^{t}_{IE}}{\partial W^{t}}}$
    \State $\mu_{q}^{t+1}=\mu_{q}^{t}-\eta^{t}\cdot\frac{\partial\mathcal{L}^{t}}{\partial \mu_{q}^{t}}=\mu_{q}^{t}-\eta^{t}\cdot\lambda\cdot\frac{\partial\mathcal{L}^{t}_{IE}}{\partial \mu_{q}^{t}}$
    \State $\sigma^{2}_{t+1}=\sigma^{2}_{t}-\eta^{t}\cdot\frac{\partial\mathcal{L}^{t}}{\partial \sigma^{2}_{t}}=\sigma^{2}_{t}-\eta^{t}\cdot\lambda\cdot\frac{\partial\mathcal{L}^{t}_{IE}}{\partial \sigma^{2}_{t}}$
    \State $\theta_{c}^{t+1}=\theta_{c}^{t}-\eta^{t}\sum^{M}_{n=1}\frac{\partial\mathcal{L}^{t}}{\partial f_{n}^{t}}\cdot\frac{\partial f_{n}^{t}}{\partial \theta_{c}^{t}}$
    \EndFor
  \end{algorithmic}
\end{algorithm}

To alleviate the computational complexity of real gradients, we assume $f_{n}, \mu_{c}, \sigma^{2}$ are three independent variables. One can refer to Appendix A for the complete derivation process. The gradients of $\mathcal{L}_{IE}(\Theta)$ with respect to $f_{n}, \mu_{c}, \sigma^{2}$ are estimated as follows:
\begin{equation}
\centering
\small{\frac{\partial\mathcal{L}_{IE}(\Theta)}{\partial f_{n}}=\frac{1}{M}\sum^{M}_{n=1}\left(\frac{f_{n}-\mu_{C(f_{n})}}{\sigma^{2}}-\frac{f_{n}}{\sigma^{2}Q}+\frac{\sum_{c=1,c\neq C(f_{n})}^{Q}exp({-\frac{1}{2\sigma^{2}Q}\|f_{n}-\mu_{c}\|^{2}_{2}})\cdot\mu_{c}}{\sigma^{2}Q\sum_{c=1,c\neq C(f_{n})}^{Q}exp({-\frac{1}{2\sigma^{2}Q}\|f_{n}-\mu_{c}\|^{2}_{2}})}\right)},
\end{equation}
\begin{equation}
\centering
\small{\frac{\partial\mathcal{L}_{IE}(\Theta)}{\partial\mu_{q}}=
\begin{cases}
\frac{1}{M}\sum^{M}_{n=1}\left(\frac{exp({-\frac{1}{2\sigma^{2}Q}\|f_{n}-\mu_{q}\|^{2}_{2}})\cdot\frac{f_{n}-\mu_{q}}{\sigma^{2}Q}}{\sum_{c=1,c\neq C(f_{n})}^{Q}exp({-\frac{1}{2\sigma^{2}Q}\|f_{n}-\mu_{c}\|^{2}_{2}})}\right),&  \ \   {q\neq C(f_{n})}\\
-\frac{1}{M}\sum^{M}_{n=1}\frac{f_{n}-\mu_{q}}{\sigma^{2}}\  \   \   \   \   \   \   \   \   \   \   \   \   \   \   \   \   \   \   \   \   \   \ ,& \   \     {q=C(f_{n})}
\end{cases}},
\end{equation}
\begin{equation}
\centering
\small{\frac{\partial\mathcal{L}_{IE}(\Theta)}{\partial \sigma^{2}}=\frac{1}{M}\sum^{M}_{n=1}\left(\frac{\sum_{c=1,c\neq C(f_{n})}^{Q}exp({-\frac{1}{2\sigma^{2}Q}\|f_{n}-\mu_{c}\|^{2}_{2}})\cdot\frac{\|f_{n}-\mu_{c}\|^{2}_{2}}{2\sigma^{4}Q}}{\sum_{c=1,c\neq C(f_{n})}^{Q}exp({-\frac{1}{2\sigma^{2}Q}\|f_{n}-\mu_{c}\|^{2}_{2}})}-\frac{\|f_{n}-\mu_{_{C(f_{n})}}\|^{2}_{2}}{2\sigma^{4}}\right)}.
\end{equation}

\section{Experiments}

The concrete implementation details are given in Section 4.1. In Section 4.2, three kinds of CNNs with different capacity are given to validate the effectiveness of our algorithm on object recognition databases (MNIST \cite{lecun1998mnist}, CIFAR10 \cite{krizhevsky2009learning} and CIFAR100 \cite{krizhevsky2009learning}). Some experiments on face recognition databases (LFW \cite{huang2014labeled} and YTF \cite{wolf2011face}) are also performed in Section 4.3.

\subsection{Implementation details}

We use the Caffe library \cite{jia2014caffe} to implement our experiments, and a speed-up parallel computing technique by two Tesla K80 GPUs is exploited. All the networks in this part are based on some existing CNNs. We partition them into three classes: the lighter, the normal and the powerful. We will refer to [L], [N] and [P] as their respective notations in the following experiments. The normal networks are shown in Table \ref{table1} and Table \ref{table5} which are inspired by \cite{liu2016large, wen2016discriminative}. Also, the powerful ones are similar to \cite{zagoruyko2016wide, he2015deep}. We adopt ReLU \cite{krizhevsky2012imagenet} as the default activation function except in Table \ref{table1} where the PReLU \cite{he2015delving} is used. The weight decay and momentum is set to 0.0005 and 0.9. Note that the mean subtraction image preprocessing is performed if not mentioned. The normally used SGD works well for the training. The lighter networks are some known structures built in Caffe library, and we comply with their original setings. In all these cases, we set $\alpha$ as $0.1$ and $Q$ as the entire inter-class distances in the mini-batch, if not specified. The joint supervision of softmax loss and IE loss is necessary to accelerate the convergence of training process. When testing, the softmax classifier is used for object recognition, and cosine similarity metric is computed to obtain the face verification accuracies. For a fair comparison, we train four kinds of models in each experiment, namely under the supervision of softmax loss, softmax loss and L-Softmax loss, softmax loss and center loss, softmax loss and IE loss. For simplicity, we refer to the four original loss names as their corresponding methods. The details of every experiment about the training setups will be presented in their respective subsections subsequently. In all the experiments, only a single model is used to achieve the final performance.

\subsection{Object recognition}
\begin{table}[h]
\centering 
\caption{Some normal CNN architectures for different benchmark datasets. Conv1.x, Conv2.x and Conv3.x denote structures that may contain multiple successive convolutional layers. Batch normalization is used in these networks.}\label{table1}
\scalebox{0.7}[0.7]{
\begin{tabular}{lllllllll}
\\
\hline
$\mathrm{\mathbf{MNIST~}(for~Fig.2)}$   &   $\mathrm{Conv0.x}$  &   $\mathrm{Conv1.x}$  &   $\mathrm{Pool1}$    &   $\mathrm{Conv2.x}$  &   $\mathrm{Pool2}$    &   $\mathrm{Conv3.x}$  &   $\mathrm{Pool3}$    &   $\mathrm{Fully~ Connected}$\\
\hline
$\mathrm{Num~ Layer}$   &   -   &   2   &   1   &   2   &   1   &   2   &   1   &   1\\
$\mathrm{Filt~ Dim}$    &   -   &   5   &   2   &   5   &   2   &   5   &   2   &   1\\
$\mathrm{Num~ Filt}$    &   -   &   32  &   -   &   64  &   -   &   128 &   -   &   2\\
$\mathrm{Stride}$   &   -   &   1   &   2   &   1   &   2   &   1   &   2   &   1\\
$\mathrm{Pad }$ &   -   &   2   &   -   &   2   &   -   &   2   &   -   &   -\\
\hline
\hline
$\mathrm{\mathbf{MNIST}}$   &   $\mathrm{Conv0.x}$  &   $\mathrm{Conv1.x}$  &   $\mathrm{Pool1}$    &   $\mathrm{Conv2.x}$  &   $\mathrm{Pool2}$    &   $\mathrm{Conv3.x}$  &   $\mathrm{Pool3}$    &   $\mathrm{Fully~ Connected}$\\
\hline
$\mathrm{Num~ Layer}$   &   1   &   3   &   1   &   3   &   1   &   3   &   1   &   1\\
$\mathrm{Filt~ Dim}$    &   3   &   3   &   2   &   3   &   2   &   3   &   2   &   1\\
$\mathrm{Num~ Filt}$    &   64  &   64  &   -   &   64  &   -   &   64  &   -   &   256\\
$\mathrm{Stride}$   &   1   &   1   &   2   &   1   &   2   &   1   &   2   &   1\\
$\mathrm{Pad }$ &   1   &   1   &   -   &   1   &   -   &   1   &   -   &   -\\
\hline
\hline
$\mathrm{\mathbf{CIFAR10}}$ &   $\mathrm{Conv0.x}$  &   $\mathrm{Conv1.x}$  &   $\mathrm{Pool1}$    &   $\mathrm{Conv2.x}$  &   $\mathrm{Pool2}$    &   $\mathrm{Conv3.x}$  &   $\mathrm{Pool3}$    &   $\mathrm{Fully~ Connected}$\\
\hline
$\mathrm{Num~ Layer}$   &   1   &   4   &   1   &   4   &   1   &   4   &   1   &   1\\
$\mathrm{Filt~ Dim}$    &   3   &   3   &   2   &   3   &   2   &   3   &   2   &   1\\
$\mathrm{Num~ Filt}$    &   64  &   64  &   -   &   96  &   -   &   128 &   -   &   256\\
$\mathrm{Stride}$   &   1   &   1   &   2   &   1   &   2   &   1   &   2   &   1\\
$\mathrm{Pad }$ &   1   &   1   &   -   &   1   &   -   &   1   &   -   &   -\\
\hline
\hline
$\mathrm{\mathbf{CIFAR100}}$    &   $\mathrm{Conv0.x}$  &   $\mathrm{Conv1.x}$  &   $\mathrm{Pool1}$    &   $\mathrm{Conv2.x}$  &   $\mathrm{Pool2}$    &   $\mathrm{Conv3.x}$  &   $\mathrm{Pool3}$    &   $\mathrm{Fully~ Connected}$\\
\hline
$\mathrm{Num~ Layer}$   &   1   &   4   &   1   &   4   &   1   &   4   &   1   &   1\\
$\mathrm{Filt~ Dim}$    &   3   &   3   &   2   &   3   &   2   &   3   &   2   &   1\\
$\mathrm{Num~ Filt}$    &   96  &   96  &   -   &   192 &   -   &   384 &   -   &   512\\
$\mathrm{Stride}$   &   1   &   1   &   2   &   1   &   2   &   1   &   2   &   1\\
$\mathrm{Pad }$ &   1   &   1   &   -   &   1   &   -   &   1   &   -   &   -\\
\hline
\end{tabular}
}
\end{table}

$\mathbf{MNIST}\    \   $ This handwritten dataset has 60,000 training images and 10,000 testing images. In this section, we use two CNNs to validate the generalization of our algorithm. One is the lighter LeNet included in Caffe library. We train it according to the default updating strategy of learning rate and parameter initialization, eventually terminate it at 12k. The normal one is depicted in Table \ref{table1}. This model is trained with the batch size of 256, and the learning rate is started from 0.01, divided by 10 at 12k and 15k iterations, eventually terminated at 20k iterations. In all these experiments, we only preprocess the images by dividing by 256 to provide them in range [0,1] as inputs. Some existing best results and the compared methods are shown in Table \ref{table2}. It is obvious that IE loss not only outperforms other DML methods under the same setings, but also among the top performance compared to other state-of-the-art methods.
\begin{table}[htp!]
\centering 
\caption{Recognition error rate ($\%$) on MNIST dataset.}\label{table2}
\scalebox{0.85}[0.85]{
\begin{tabular}{ lc }
\\
\hline
$\mathrm{Method}$~~~~~~~~~~~~~~~~~~~~~~~~~~~~~~~~~~~~~~~~~~~~~~~~~~~~~~~~~~~~~~~~~~~~~~~~~~~~    &   $\mathrm{Error~Rate}$ (\%)\\
\hline\hline
$\mathrm{DropConnect}$ \cite{wan2013regularization} &   0.57\\

$\mathrm{CNN}$ \cite{jarrett2009best}   &   0.53\\

$\mathrm{Maxout}$ \cite{goodfellow2013maxout}   &   0.45\\

$\mathrm{DSN}$ \cite{lee2015deeply} &   0.39\\

R-CNN \cite{liang2015recurrent} &   $\mathbf{0.31}$ \\

$\mathrm{GenPool}$ \cite{lee2016generalizing}   &   $\mathbf{0.31}$\\
\hline
\hline
$\mathrm{Softmax~[L]}$  &   0.83\\

L-Softmax~[L]   &   0.74\\

$\mathrm{Center~[L]}$   &   0.76\\

$\mathrm{IE~[L]}$   &   $\mathbf{0.49}$\\
\hline
\hline
$\mathrm{Softmax~[N]}$  &   0.61\\

L-Softmax~[N]   &   0.47\\

$\mathrm{Center~[N]}$   &   0.58\\

$\mathrm{IE~[N]}$   &   $\mathbf{0.31}$\\
\hline
\hline
\end{tabular}
}
\end{table}

$\mathbf{CIFAR10}\  \   $ This dataset has 10 classes of objects with 50k for training and 10k for testing. The experiments on three CNNs are carried out here. The lighter one is the Cifar10 network built in Caffe library. The updating strategy and initialization of parameters follow the original settings. The normal one is depicted in Table \ref{table1}. We start with a learning rate of 0.01, divide it by 10 at 10k and 17k iterations, and eventually terminate it at 22k iterations. Simple mean/std normalization and horizontal flips are used to preprocess the dataset. The powerful one is WRN-28-10 as illustrated in \cite{zagoruyko2016wide}, with some differences. The WRN-28-10 network is said to achieve a comparable accuracy with more than 1000 layers raw ResNet \cite{he2015deep} on CIFAR10. To speed up the training process, we fine-tune the other three compared DML methods from the softmax baseline model. In this experiment, the dataset is preprocessed by global contrast normalization and mean/std normalization. We follow the standard data augmentation \cite{liu2016large} for training, and the batch size is 128. The results are listed in Table \ref{table3}. We can observe that our method always achieves the best performance among the four compared DML methods regardless of the size of CNNs.
\begin{table}[htp!]
\centering 
\caption{Recognition error rate ($\%$) on CIFAR10 dataset.}\label{table3}
\scalebox{0.85}[0.85]{
\begin{tabular}{ lc }
\\
\hline
$\mathrm{Method}$~~~~~~~~~~~~~~~~~~~~~~~~~~~~~~~~~~~~~~~~~~~~~~~~~~~~~~~~~~~~~~~~~~~~~~~~~~~~   &   $\mathrm{Error~Rate}$ (\%)\\
\hline\hline
$\mathrm{Maxout}$ \cite{goodfellow2013maxout}   &   11.68\\

$\mathrm{DSN}$ \cite{lee2015deeply} &   9.69\\

$\mathrm{DropConnect}$ \cite{wan2013regularization} &   9.41\\

All-CNN \cite{springenberg2014striving} &   9.08\\

R-CNN \cite{liang2015recurrent} &   8.69\\

$\mathrm{GenPool}$ \cite{lee2016generalizing}   &   $\mathbf{7.62}$\\
\hline
\hline
$\mathrm{Softmax~[L]}$  &   21.88\\

L-Softmax~[L]   &   -\\

$\mathrm{Center~[L]}$   &   19.40\\

$\mathrm{IE~[L]}$   &   $\mathbf{18.98}$\\
\hline
\hline
$\mathrm{Softmax~[N]}$  &   11.56\\

L-Softmax~[N]   &   9.59\\

$\mathrm{Center~[N]}$   &   10.25\\

$\mathrm{IE~[N]}$   &   $\mathbf{8.77}$\\
\hline
\hline
$\mathrm{Softmax~[P]}$  &   6.59\\

L-Softmax~[P]   &   6.46\\

$\mathrm{Center~[P]}$   &   6.17\\

$\mathrm{IE~[P]}$   &   $\mathbf{5.97}$\\
\hline
\end{tabular}
}
\end{table}

$\mathbf{CIFAR100}\  \   $ The final part of this section, we will verify the effectiveness of IE loss on CIFAR100 dataset. This dataset is just like the CIFAR10, except it has 100 classes containing 600 images  per class, where 500 for training and 100 for testing. The 100 classes in CIFAR100 are grouped into 20 superclasses. Each image comes with a ``fine'' label (the class to which it belongs) and a ``coarse'' label (the superclass to which it belongs). We use the former protocol here. By convention, the normal network is shown in Table  \ref{table1}, and the powerful one is WRN-28-10. Also, the training strategy is the same as which described in CIFAR10. For the powerful WRN-28-10, we fine-tune the other three compared DML methods from the softmax baseline model. Differently, to better inspect the effectiveness of the compared methods with the capacity of networks growing, we preprocess the dataset in the same way on the normal and powerful networks, only by simple mean/std normalization and horizontal flips to augment data. In Table \ref{table4}, we can clearly find that our method consistently performs better than other compared approaches.
\begin{table}[htp!]
\centering 
\caption{Recognition error rate ($\%$) on CIFAR100 dataset.}\label{table4}
\scalebox{0.9}[0.9]{
\begin{tabular}{ lc }
\\
\hline
\small{$\mathrm{Method}$}    ~~~~~~~~~~~~~~~~~~~~~~~~~~~~~~~~~~~~~~~~~~~~~~~~~~~~~~~~~~~~~~~~~~~~~~~~~~~~&   $\mathrm{Error~Rate}$ (\%) \\

\hline\hline
$\mathrm{Maxout}$ \cite{goodfellow2013maxout}   &   38.57\\

$\mathrm{DSN}$ \cite{lee2015deeply} &   34.57\\

All-CNN \cite{springenberg2014striving} &   33.71\\

R-CNN \cite{liang2015recurrent} &   $\mathbf{31.75}$\\

\hline
\hline
$\mathrm{Softmax~[N]}$  &   33.31\\

L-Softmax~[N]   &   30.79\\

$\mathrm{Center~[N]}$   &   29.39\\

$\mathrm{IE~[N]}$   &   $\mathbf{28.42}$\\
\hline
\hline
$\mathrm{Softmax~[P]}$  &   27.06\\

L-Softmax~[P]   &   26.21\\

$\mathrm{Center~[P]}$   &   26.15\\

$\mathrm{IE~[P]}$   &   $\mathbf{25.32}$\\
\hline
\end{tabular}
}
\end{table}

\begin{figure}[H]
 \centering
 \subfigure[]{
 \label{a}
 \includegraphics[height=6cm,width=14cm]{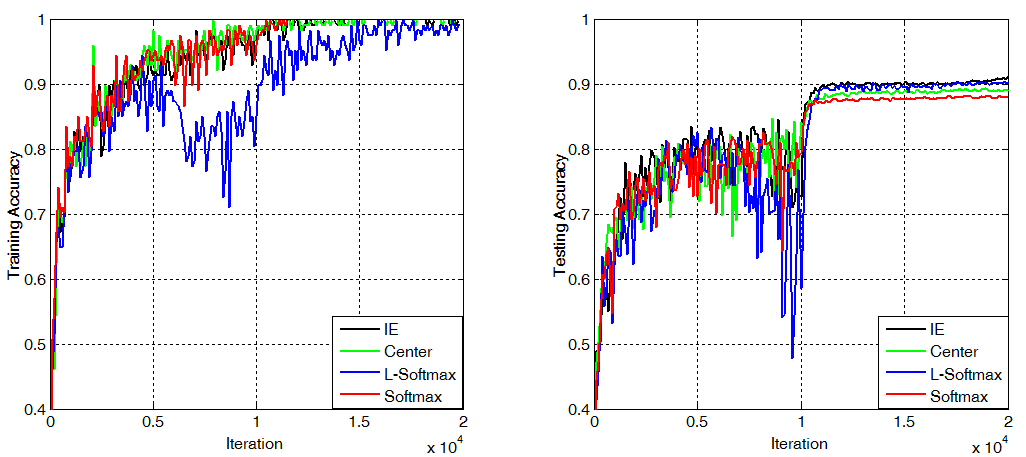}}
 \subfigure[]{
 \label {b}
 \includegraphics[height=6cm,width=14cm]{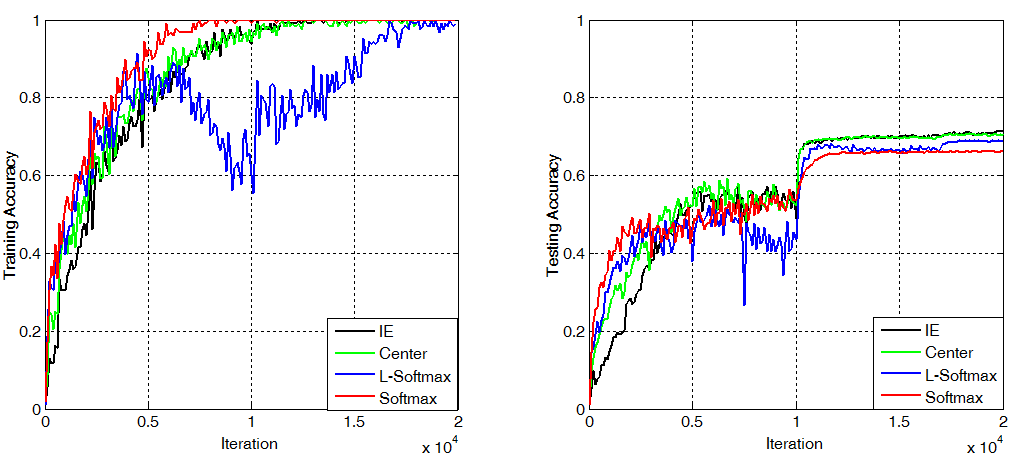}}
 \caption{Accuracy vs. iteration curves using the normal networks on (a) CIFAR10 dataset and (b) CIFAR100 dataset.}\label{fig3}
\end{figure}

From the results presented above, one can find that our IE loss always achieves the best results among the four compared DML methods on three object recognition datasets. Specifically, the performance of center loss and L-Softmax loss fluctuates significantly with different network structures. In Fig.~\ref{fig3}, the training and testing process on CIFAR10 and CIFAR100 with the normal CNNs are displayed. It can be seen that the convergence rate of our IE loss is comparable with other compared loss functions, avoiding the notoriously slow convergence of triplet loss. Considering the performance gap between training and testing, one can observe that IE loss can mitigate the serious overfitting of softmax loss and the difficult convergence of L-Softmax loss. The testing accuracies of our method about different $\lambda$ and $\alpha$, and the best settings of them on the normal networks are shown in Appendix B

\subsection{Face verification}

\begin{figure}[htp!]\centering
 \includegraphics[height=6.3cm,width=13.5cm]
 {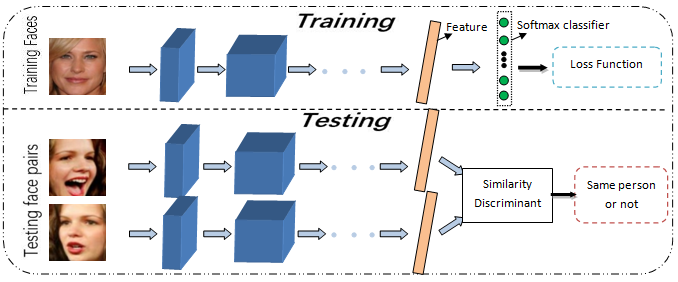}
 \caption{The general pipeline for face verification in this paper, where classifier loss function is used to train and similarity discriminant is used to obtain the final verification accuracy.}\label{fig4}
\end{figure}

\begin{table}[h]
\centering
\tabcolsep 12.5pt
\caption{The normal ResNet architecture used for face verification. Resblock is the classical Residual unit which consists of two consecutive convolutional layers and a unit mapping.}\label{table5}
\scalebox{0.75}[0.75]{
\begin{tabular}{ lccccc }
\\
\hline
$\mathrm{Layer}$    &   $\mathrm{Type}$ &   $\mathrm{Filter~Size/Stride}$   &   $\mathrm{Output~Size}$  &   $\mathrm{Depth}$    &   $\mathrm{Params}$\\
\hline
$\mathrm{Conv0}$    &   convolution &   $3\times3/1$    &   $110\times94\times32$   &   1   &   0.86K\\
$\mathrm{Conv1}$    &   convolution &   $3\times3/1$    &   $108\times92\times64$   &   1   &   18K\\
$\mathrm{Pool1}$    &   max pooling &   $2\times2/2$    &   $54\times46\times64$    &   0   &   -\\
$\mathrm{Resblock1}$    &   convolution &   $3\times3/1$    &   $54\times46\times64$    &   2   &   73K\\
$\mathrm{Conv2}$    &   convolution &   $3\times3/1$    &   $52\times44\times128$   &   1   &   73K\\			
$\mathrm{Pool2}$    &   max pooling &   $2\times2/2$    &   $26\times22\times128$   &   0   &   -\\			
$\mathrm{Resblock2}$    &   convolution &   $3\times3/1$    &   $26\times22\times128$   &   2   &   294K\\
$\mathrm{Resblock3}$    &   convolution &   $3\times3/1$    &   $26\times22\times128$   &   2   &   294K\\
$\mathrm{Conv3}$    &   convolution &   $3\times3/1$    &   $24\times20\times256$   &   1   &   294K\\
$\mathrm{Pool3}$    &   max pooling &   $2\times2/2$    &   $12\times10\times256$   &   0   &   -\\
$\mathrm{Resblock4}$    &   convolution &   $3\times3/1$    &   $12\times10\times256$   &   2   &   1179K\\
$\mathrm{Resblock5}$    &   convolution &   $3\times3/1$    &   $12\times10\times256$   &   2   &   1179K\\
$\mathrm{Resblock6}$    &   convolution &   $3\times3/1$    &   $12\times10\times256$   &   2   &   1179K\\
$\mathrm{Resblock7}$    &   convolution &   $3\times3/1$    &   $12\times10\times256$   &   2   &   1179K\\
$\mathrm{Resblock8}$    &   convolution &   $3\times3/1$    &   $12\times10\times256$   &   2   &   1179K\\
$\mathrm{Conv4}$    &   convolution &   $3\times3/1$    &   $10\times8\times512$    &   1   &   1179K\\
$\mathrm{Pool4}$    &   max pooling &   $2\times2/2$    &   $5\times4\times512$ &   0   &   -\\
$\mathrm{Resblock9}$    &   convolution &   $3\times3/1$    &   $5\times4\times512$ &   2   &   4718K\\
$\mathrm{Resblock10}$   &   convolution &   $3\times3/1$    &   $5\times4\times512$ &   2   &   4718K\\
$\mathrm{Resblock11}$   &   convolution &   $3\times3/1$    &   $5\times4\times512$ &   2   &   4718K\\
$\mathrm{Fc5}$  &   fully connection    &   -   &   $1\times1\times512$ &   1   &   5242K\\
\hline
\end{tabular}
}
\end{table}

Different from object recognition, face verification is to compute the feature similarity of two images, and threshold comparison is exploited to decide whether the same person or not. Specifically, we use softmax classifier and metric loss functions to jointly supervise the training process, and the cosine similarity of two features is used to obtain the testing accuracy (Fig.~\ref{fig4}). In this section, we evaluate our approach for face verification on LFW and YTF datasets. These two face datasets are the recognized benchmarks for face image and video, respectively. We use the publicly available CASIA-WebFace \cite{yi2014learning} as the training set, which originally has 494,414 labeled face images from 10,575 individuals. After removing the images failing to detect and mislabeled, the resulting dataset for our training is just over 430K images. The cropped faces of all images are detected by \cite{wu2017funnel}, and 5 facial landmarks are labeled to globally align the face images by similarity transformation \cite{zhang2014coarse}. The normal network is depicted in Table \ref{table5}, which is a reduced version of ResNet \cite{he2015deep} with 27 convolutional layers. The input faces are cropped to $112\times96$ RGB images, and the batch size is 256. Besides, the images are normalized by subtracting the mean image and dividing by 128. We start the training with a learning rate of 0.1, and divide it by 10 at 16K, 24K iterations, then terminate it at 28K iterations. For face images, we find that using wider ResNet with fewer layers like WRN-28-10 does not bring so many benefits, and accompanied by rapidly growing memory space. So we decide to widen the network listed in Table \ref{table5} to obtain the powerful one. Specifically, we widen all the convolutional layers between Conv1 and Conv4 with a widening factor 2. When testing, we extract the features from both the frontal face and its mirror image, and merge the two features by element-wise summation.  All the evaluations are based on the similarity scores of image pairs, which are computed by the cosine similarity of two representations after PCA.

\begin{figure}[h]\centering
\subfigure[]{
\label{a}
 \includegraphics[height=5.3cm,width=6.6cm]
 {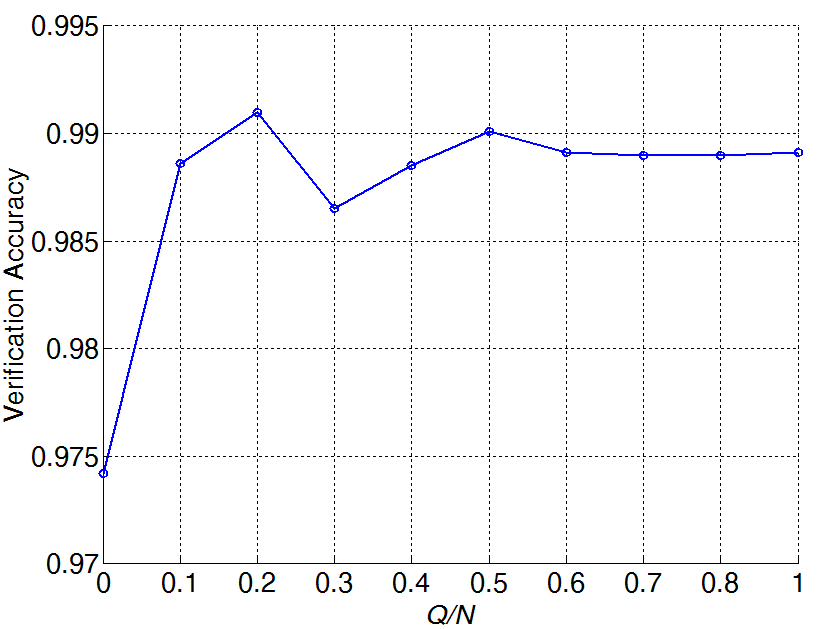}}
 \subfigure[]{
 \label {b}
 \includegraphics[height=5.3cm,width=6.6cm]
 {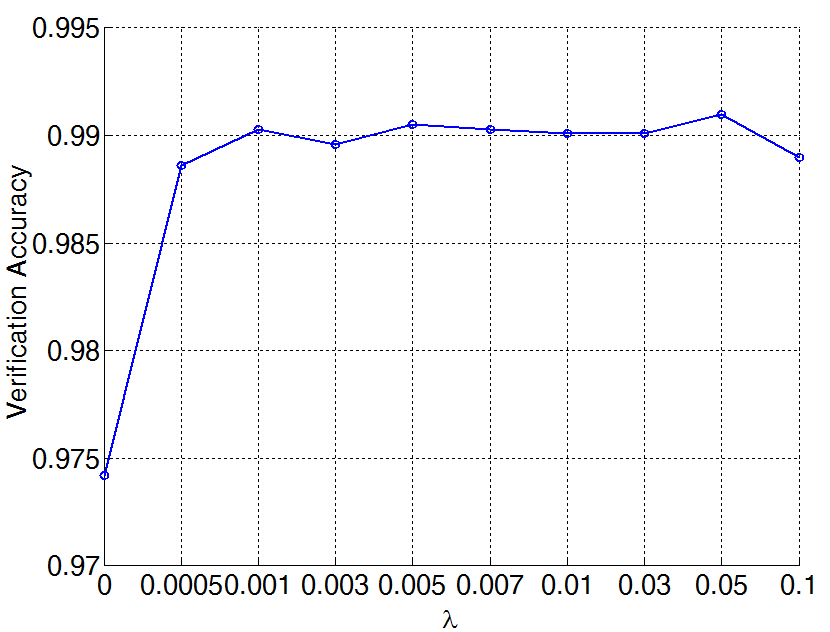}}
 \caption{(a) Verification accuracies of IE loss with different $Q/N$ on LFW using the normal network, where $N$ is the number of inter-class distances regarding a sample in a mini-batch. (b) Face verification accuracies of IE Loss on LFW with different $\lambda$ using the normal network.}\label{fig5}
\end{figure}

Considering the difference from previous experiments, we select $Q$ as the first $20\%$ inter-class distances in every mini-batch to calculate the objective here. The reason is that some datasets like CASIA-WebFace have too many subjects, most of the inter-class distances tend to be very large in our method, thus leading to the difficult convergence of training process. Fig.~\ref{fig5}a shows the verification accuracies on LFW with $Q$ ranging from $0$ to $100\%$ of the number of inter-class distances. The importance of choosing a proper $Q$ is displayed clearly. Here, we regard the case when $Q=0$ as the original softmax method.

$\mathbf{LFW}\  \   $This dataset contains 13,233 face images of 5,749 different identities from the Internet, with large variations in pose, expression and illumination. For comparison purpose, algorithms typically report the mean face verification accuracies and the ROC curves on 6000 given face pairs, following the standard protocol of unrestricted with labeled outside data \cite{huang2014labeled}. According to previous experience, we find that a properly chosen $\lambda$ which balances the weight between softmax loss and IE loss can improve the performance. So we experiment our method across a wide range of $\lambda$ from 0 to 0.1 to select the best setting. The results on LFW are shown in Fig.~\ref{fig5}b. It can be seen that IE loss is stable with different $\lambda$, and the best setting is 0.05.

\begin{figure}[h]
 \centering
 \subfigure[]{
 \label{a}
 \includegraphics[height=5.35cm,width=6.6cm]
 {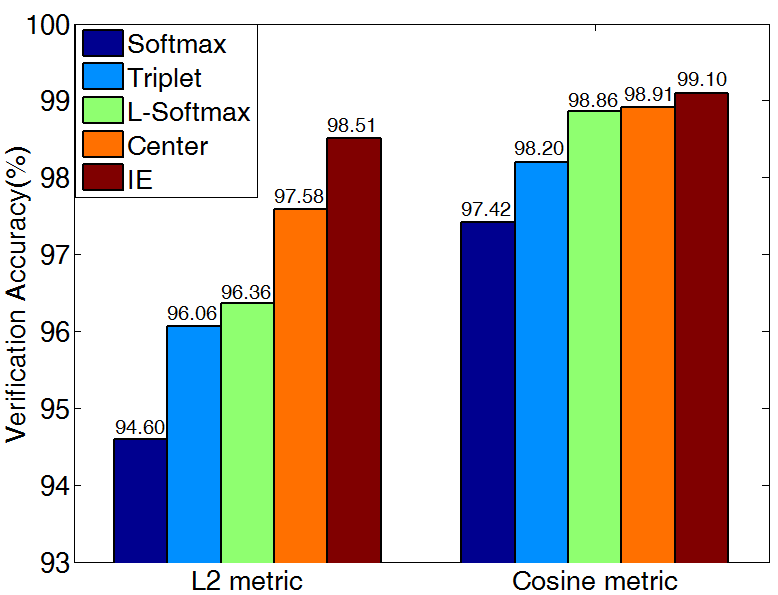}}
 \subfigure[]{
 \label {b}
 \includegraphics[height=5.4cm,width=6.6cm]
 {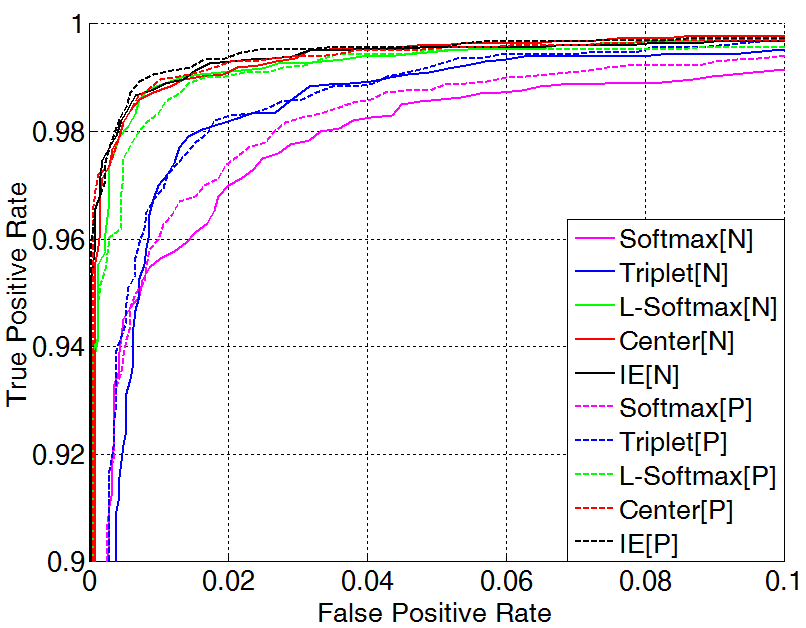}}
 \caption{(a)Verification accuracies of compared loss functions with two different similarity metrics on LFW using the normal network. (b) ROC curves of five compared loss functions on LFW.}\label{fig6}
\end{figure}

Fig.~\ref{fig6}a illustrates the verification accuracies of five loss functions with two different similarity metrics for testing. The results show that cosine similarity is more suitable than L2 similarity for our feature representations. Obviously, our method is robust to both cases, and always achieves the best performance.

\begin{figure}[htp!]
 \centering
 \subfigure[]{
 \label{a}
 \includegraphics[height=5.4cm,width=6.6cm]
 {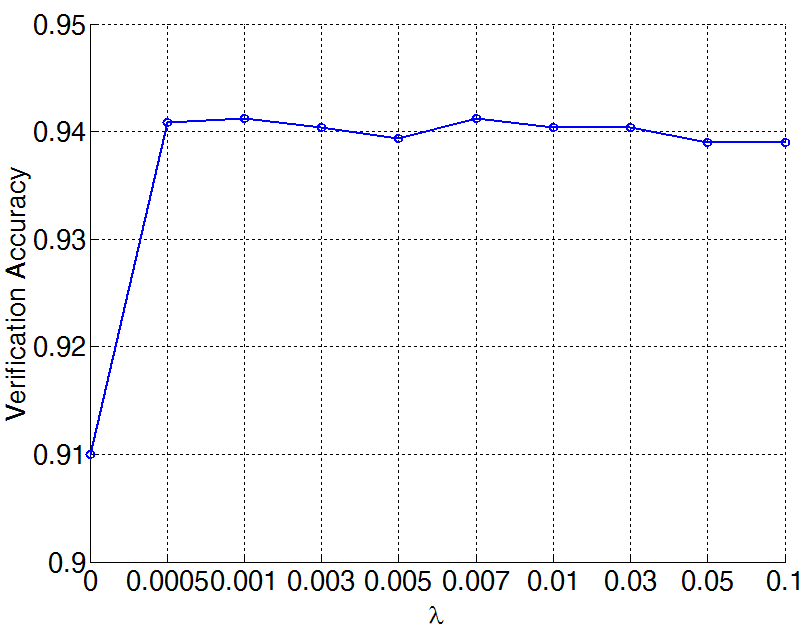}}
 \subfigure[]{
 \label {b}
 \includegraphics[height=5.4cm,width=6.6cm]
 {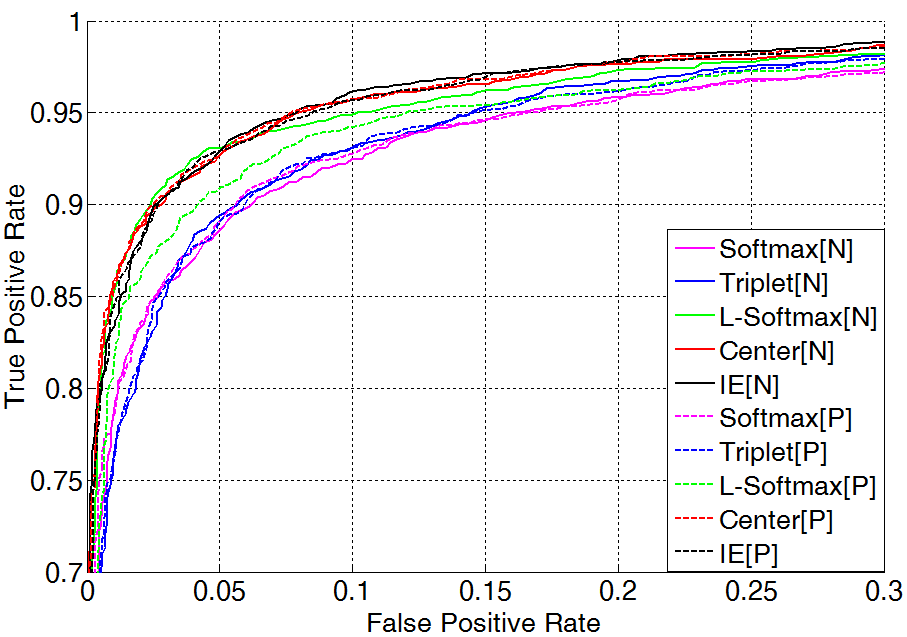}}
 \caption{(a) Face verification accuracies of IE Loss on YTF with different $\lambda$ using the normal ResNet. (b) ROC curves of five compared loss functions on YTF.}\label{fig7}
\end{figure}

$\mathbf{YTF}\  \   $This dataset consists of 3,425 videos from 1,595 different people, with an average of 2.15 videos for everyone. Besides, the average length of a video clip is 181.3 frames, with each clip duration varying from 48 frames to 6,070 frames. Just as the experiments on LFW, we report the results on 5,000 video pairs in Table \ref{table6}, according to the unrestricted protocol with labeled outside data in \cite{wolf2011face}. Also, Fig.~\ref{fig7} shows the accuracy of IE loss in regard to different $\lambda$ ranging from 0 to 0.1 and the ROC curves of five compared loss functions.

\begin{figure}[htp!]
 \centering
 \subfigure[Samples of CIFAR100]{
 \label{a}
 \includegraphics[height=3cm,width=13.6cm]
 {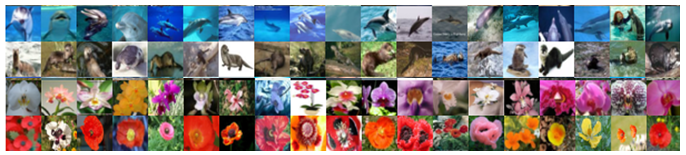}~~~~~~~~~~}
 \subfigure[Face images in LFW]{
 \label {b}
 \includegraphics[height=2.2cm,width=13.6cm]
 {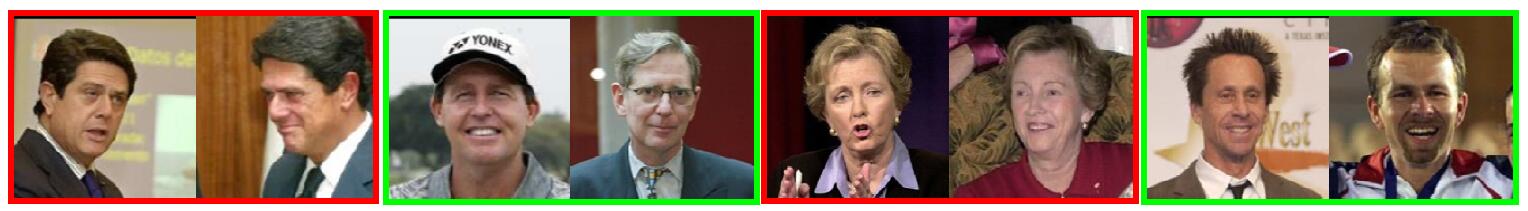}~~~~~~~~~~}
 \caption{Some examples of the datasets in our experiments. The image pairs in red are those positive pairs that our method succeeds to recognize, while the softmax method fails. Likewise, the green ones are some negative pairs.}\label{fig8}
\end{figure}

From the verification results in Table \ref{table6} and ROC curves on these two datasets, we can find that the performance on the powerful network is consistently superior to which on the normal one except the L-Softmax loss. IE loss is always outstanding in the five loss functions under a small training dataset of CASIA-WebFace, and competitive with the state-of-the-art methods using larger training datasets or model ensemble. Noticeably, the results of triplet loss and L-Softmax loss are not satisfactory, and there exhibits a large margin of triplet loss compared to the results in \cite{schroff2015facenet}. This convincingly demonstrates the difficult convergence and big data dependence of triplet loss. We conjecture that maybe the performance of our method can be improved considerably if a larger training set or a more powerful network is used. Anyway, the excellent performance undoubtedly verify the great generalization of IE loss. The visualization of some datasets is shown in Fig.~\ref{fig8}.

\begin{table}[htp!]
\centering \tabcolsep 12pt
\caption{Face verification performance ($\%$) on LFW and YTF datasets.}\label{table6}
\scalebox{0.61}[0.65]{
\begin{tabular}{ lccccc }
\\
\hline
$\mathrm{Method}$   &   $\mathrm{Points~for~ Alig.}$    &   $\mathrm{Outside~ Data}$    &  $\mathrm{Networks}$ &   $\mathrm{Acc.~ on~ LFW}$ (\%)   &   $\mathrm{Acc.~ on~ YTF}$ (\%)\\
\hline\hline
High-dim LBP \cite{chen2013blessing}    &   27  &   100K    &   -   &   $95.17$ &   -\\
$\mathrm{DeepFace}$\cite{taigman2014deepface}   &   73  &   4M  &   3   &   97.35   &   91.40\\
$\mathrm{Gaussian~ Face}$ \cite{lu2014surpassing}   &   -   &   20K &   1   &   98.52   &   -\\
$\mathrm{DeepID}$ \cite{sun2014deep}    &   5   &   200K    &   1   &   97.45   &   -\\

DeepID-2+ \cite{sun2015deeply}  &   18  &   300K    &   25  &   99.47   &   93.20\\

$\mathrm{FaceNet}$ \cite{schroff2015facenet}    &   -   &   200M    &   1   &   $\mathbf{99.63}$    &   $\mathbf{95.10}$\\

$\mathrm{DCNN}$ \cite{chen2016unconstrained}    &   7   &   490K    &   1   &   97.45   &   -\\

CASIA-WebFace \cite{yi2014learning} &   2   &   490K    &   1   &   97.73   &   90.60\\
\hline
\hline
$\mathrm{Softmax~[N]}$  &   5   &   430K    &   1   &   97.42   &   91.52\\

$\mathrm{Triplet~Loss~[N]}$ &   5   &   430K    &   1   &   98.20   &   92.16\\

L-Softmax~[N]  &   5   &   430K    &   1   &   98.86   &   $\mathbf{94.14}$\\

$\mathrm{Center~[N]}$   &   5   &   430K    &   1   &   98.91   &   93.80\\

$\mathrm{IE~[N]}$   &   5   &    430K   &   1   &   $\mathbf{99.10}$  &  94.12\\
\hline
\hline
$\mathrm{Softmax~[P]}$  &   5   &   430K    &   1   &   97.73   &   92.42\\

$\mathrm{Triplet~Loss~[P]}$ &   5   &   430K    &   1   &   98.23   &   91.98\\

L-Softmax~[P]   &   5   &   430K    &   1   &   98.67   &   92.66\\

$\mathrm{Center~[P]}$   &   5   &   430K    &   1   &   99.01   &   $\mathbf{94.12}$\\

$\mathrm{IE~[P]}$   &   5   &   430K    &   1   &   $\mathbf{99.15}$   &    $\mathbf{94.12}$\\
\hline
\end{tabular}
}
\end{table}


\section{Conclusion and future work}

 In this paper, we propose a powerful and replicable DML method, which enforces the mean inter-class distance larger than the intra-class distance with a margin, to enhance the discriminability of the deeply learned features in object recognition and face verification. Extensive experiments on several public datasets have convincingly demonstrated the effectiveness of our method. The results also exhibit the excellent generalization of IE loss in various size of CNNs. Instead of requiring a superior neighborhood sampling strategy, our approach only uses mini-batch based SGD to conduct the experiments, avoiding the exponentially increased computational complexity of image pairs or triplets. Maybe a better hard sample mining strategy could improve the performance further. Inspired by the outstanding performance of IE loss in object recognition and face recognition, we will explore its extension in the case where the swarm intelligent methods are exploited to optimize the clustering algorithm \cite{zeng2016deep, zeng2018facial} in the following work. In the future, we will delve into DML to explore its extensive applications to other tasks.

\section*{Acknowledgements}
The authors would like to thank Kun Shang, Mengya Zhang, Ruipeng Shen and Wenjuan Li for their helpful advices. This research was supported by the National Science Foundation of China.

\section*{Appendix A}
In this section, we concretely describe the deduction of gradient formulas $(9)\sim(11)$ listed in Section 3.2. First, we rewrite Eq.(6) as follows:
\begin{equation}
\mathcal{L}= \frac{1}{M}\sum^{M}_{n=1}\left\{-log\frac{exp({-\frac{1}{2\sigma^{2}}\|f_{n}-\mu_{C(f_{n})}\|^{2}_{2}-\alpha})}{\sum_{c=1,c\neq C(f_{n})}^{Q}exp({-\frac{1}{2\sigma^{2}Q}\|f_{n}-\mu_{c}\|^{2}_{2}})}\right\}_{+}.
\tag{A.1}
\end{equation}

\noindent We need to compute the gradient formulas of $\mathcal{L}$ with respect to $f_{n}, \mu_{c}$ and $\sigma^{2}$.  Note that directly computing the real gradients of them leads to costly computational complexity in training. So we will consider $f_{n}, \mu_{c}$ and $\sigma^{2}$ as three independent variables. If the value in $\{\cdot\}$ is positive, then
\begin{eqnarray*}
\frac{\partial\mathcal{L}}{\partial f_{n}}&=& -\frac{1}{M}\cdot\frac{\partial}{\partial f_{n}}\left(\sum^{M}_{n=1}log\frac{exp({-\frac{1}{2\sigma^{2}}\|f_{n}-\mu_{C(f_{n})}\|^{2}_{2}-\alpha})}{\sum_{c=1,c\neq C(f_{n})}^{Q}exp({-\frac{1}{2\sigma^{2}Q}\|f_{n}-\mu_{c}\|^{2}_{2}})}\right) \\
 &=& \frac{1}{M}\cdot\frac{\partial}{\partial f_{n}}\left(\frac{\|f_{n}-\mu_{C(f_{n})}\|^{2}_{2}}{2\sigma^{2}}+\alpha+\log{\sum_{c=1,c\neq C(f_{n})}^{Q}exp({-\frac{1}{2\sigma^{2}Q}\|f_{n}-\mu_{c}\|^{2}_{2}})}\right)
 \end{eqnarray*}
 \begin{equation}
 ~~~~~~~~~~=~~~ \frac{1}{M}\sum^{M}_{n=1}\left(\frac{f_{n}-\mu_{C(f_{n})}}{\sigma^{2}}-\frac{f_{n}}{\sigma^{2}Q}+\frac{\sum_{c=1,c\neq C(f_{n})}^{Q}exp({-\frac{1}{2\sigma^{2}Q}\|f_{n}-\mu_{c}\|^{2}_{2}})\cdot\mu_{c}}{\sigma^{2}Q\sum_{c=1,c\neq C(f_{n})}^{Q}exp({-\frac{1}{2\sigma^{2}Q}\|f_{n}-\mu_{c}\|^{2}_{2}})}\right).
 \tag{A.2}
\end{equation}

\begin{equation}
\frac{\partial\mathcal{L}}{\partial\mu_{q}}= \frac{1}{M}\cdot\frac{\partial}{\partial \mu_{q}}\left(\frac{\|f_{n}-\mu_{C(f_{n})}\|^{2}_{2}}{2\sigma^{2}}+\alpha+\log{\sum_{c=1,c\neq C(f_{n})}^{Q}exp({-\frac{1}{2\sigma^{2}Q}\|f_{n}-\mu_{c}\|^{2}_{2}})}\right).
\tag{A.3}
\end{equation}

\noindent When $q\neq C(f_{n})$, we have
\begin{equation}
\frac{\partial\mathcal{L}}{\partial\mu_{q}}= \frac{1}{M}\sum^{M}_{n=1}\left(\frac{exp({-\frac{1}{2\sigma^{2}Q}\|f_{n}-\mu_{q}\|^{2}_{2}})\cdot\frac{f_{n}-\mu_{q}}{\sigma^{2}Q}}{\sum_{c=1,c\neq C(f_{n})}^{Q}exp({-\frac{1}{2\sigma^{2}Q}\|f_{n}-\mu_{c}\|^{2}_{2}})}\right).
\tag{A.4}
\end{equation}

\noindent When $q= C(f_{n})$, we have
\begin{equation}
\frac{\partial\mathcal{L}}{\partial\mu_{q}}=-\frac{1}{M}\sum^{M}_{n=1}\frac{f_{n}-\mu_{q}}{\sigma^{2}}.
\tag{A.5}
\end{equation}

\begin{eqnarray*}
\frac{\partial\mathcal{L}}{\partial \sigma^{2}}&=&\frac{1}{M}\cdot\frac{\partial}{\partial \sigma^{2}}\left(\frac{\|f_{n}-\mu_{C(f_{n})}\|^{2}_{2}}{2\sigma^{2}}+\alpha+\log{\sum_{c=1,c\neq C(f_{n})}^{Q}exp({-\frac{1}{2\sigma^{2}Q}\|f_{n}-\mu_{c}\|^{2}_{2}})}\right)
\end{eqnarray*}
\begin{equation}
 ~~~~~~~~~~=~~~ \frac{1}{M}\sum^{M}_{n=1}\left(\frac{\sum_{c=1,c\neq C(f_{n})}^{Q}exp({-\frac{1}{2\sigma^{2}Q}\|f_{n}-\mu_{c}\|^{2}_{2}})\cdot\frac{\|f_{n}-\mu_{c}\|^{2}_{2}}{2\sigma^{4}Q}}{\sum_{c=1,c\neq C(f_{n})}^{Q}exp({-\frac{1}{2\sigma^{2}Q}\|f_{n}-\mu_{c}\|^{2}_{2}})}-\frac{\|f_{n}-\mu_{C(f_{n})}\|^{2}_{2}}{2\sigma^{4}}\right).
\tag{A.6}
\end{equation}

\section*{Appendix B}

\setcounter{table}{0}
\renewcommand{\thetable}{B.\arabic{table}}

\begin{table}[H]
\centering
\caption{The recognition accuracy of IE loss on MNIST regarding different value of $\lambda$ and $\alpha$ respectively with (a) LeNet built in Caffe library and (b) MNIST network depicted in Tab.1.}\label{tableb1}
\subtable[]{
\scalebox{0.7}[0.7]{
	\begin{tabular}{cccc}
		\hline
	  $\lambda$  &   accuracy  &   $\alpha$  &    accuracy\\		
		\hline
		0.110   &   0.9939    &   0.01    &  0.9945 \\
		0.115   &   0.9936    &   0.03   &  0.9939\\
		0.120   &   0.9940    &   0.05   &  0.9938 \\
		0.125   &   0.9949    &   0.07   &   0.9943\\
		0.130   &   0.9944    &  \textbf{0.10}   &   \textbf{0.9951}   \\
        0.135   &   0.9940     &   0.15   &   0.9950   \\
        0.140   &   0.9938    &   0.20   &   0.9936   \\
        0.150   & 0.9937      &   0.25   &   0.9945   \\
        0.170   &0.9935       &   0.30   &   0.9943   \\
        0.190   &0.9938   &   0.35   &   0.9945   \\
        0.210   &0.9943   &   \textbf{0.40}   & \textbf{0.9951}   \\
        0.230   &0.9945   &   0.45   &   0.9938   \\
        0.250   & 0.9945  &   0.50   &   0.9942   \\
        0.270   &0.9945   &   0.55   &   0.9947   \\
        0.290& 0.9946     &   0.60   &   0.9941   \\
        0.310& 0.9944     &   0.65   &   0.9937   \\
        0.330& 0.9945     &   0.70   &   0.9948   \\
        0.350& 0.9943     &   0.75   &   0.9946   \\
        0.370&0.9947      &   0.80   &   0.9942   \\
        0.390& 0.9938     &   0.85   &   0.9945   \\
        0.410& 0.9943     &   0.90   &   0.9940   \\
       \textbf{0.430}&\textbf{0.9951}      &   0.95   &   0.9940   \\
        0.450& 0.9945     &   1.00   &   0.9938   \\
        0.470& 0.9945     &      &      \\
        0.500& 0.9948     &      &      \\
        0.550& 0.9946     &      &      \\
        0.600& 0.9943     &     &      \\
        0.650& 0.9942     &     &      \\
        0.700& 0.9937 &     &      \\
        0.750& 0.9945&     &      \\
        0.800& 0.9942 &     &      \\
        0.850& 0.9944  &     &      \\
        0.900& 0.9941  &     &      \\
        0.950& 0.9949 &     &      \\
        1.000& 0.9938&     &      \\
		\hline																
	\end{tabular}}}
\subtable[]{
\scalebox{0.7}[0.7]{
	\begin{tabular}{cccc}
		\hline
	  $\lambda$  &   accuracy  &   $\alpha$  &    accuracy \\		
		\hline
		0.001   &   0.9964    &   0.01    &  0.9961 \\
		0.004   &   0.9958    &   0.03   &  0.9965\\
		0.007   &   0.9952    &   0.05   &  0.9962 \\
		0.010   &   0.9963    &   0.07   &   0.9967\\
		0.030   &   0.9961    &   0.09   &   0.9962   \\
        0.050   &   0.9962     &  \textbf{0.10}   & \textbf{0.9969}   \\
        0.070   &   0.9958    &   0.13   &   0.9956   \\
        0.090   & 0.9961      &   0.15   &   0.9959   \\
        0.110   &0.9961       &   0.18   &   0.9958   \\
        0.130   &0.9961   &   0.21   &   0.9966   \\
        \textbf{0.150}   &\textbf{0.9969}   &   0.24   &   0.9967   \\
        0.170   &0.9963   &   0.27   &   0.9963   \\
        0.190   & 0.9955  &   0.30     & 0.9958  \\
        0.210   &0.9963   &          &   \\
        0.230   & 0.9960     &    &    \\
        0.250   & 0.9957     &    &    \\
        0.270   & 0.9959     &    &    \\
		\hline																
	\end{tabular}}
}
\end{table}

\begin{table}[H]
\centering
\caption{The recognition accuracy of IE loss on CIFAR10 regarding different value of $\lambda$ and $\alpha$ respectively with (a) CIFAR10 built in Caffe library and (b) CIFAR10 network depicted in Tab.1.}\label{tableb2}
\subtable[]{
\scalebox{0.7}[0.7]{
	\begin{tabular}{cccc}
		\hline
	  $\lambda$  &   accuracy  &   $\alpha$  &    accuracy\\		
		\hline
		0.001   &   0.8028    &   0.001    &  0.8057 \\
		0.004   &   0.8054    &   0.005   &  0.8018\\
		0.008   &   0.8064    &   0.010  &  0.8068 \\
		0.010   &   0.8063    &   0.050   &   0.8029\\
		0.040   &   0.8011    &   0.100   &    0.8093  \\
        0.080   &   0.7950     &   0.150   &   0.8032   \\
        0.100   &   0.8033    &   0.200   &   0.7972   \\
        0.130   & 0.8012      &   0.250   &   0.7989   \\
        0.160   &0.8064       &   0.300   &   0.7996   \\
        0.190   &0.7959   &   0.350   &   0.8059   \\
        0.210   &0.7998   &   \textbf{0.40}   & \textbf{0.8102}   \\
        0.240   &0.8002   &   0.450   &   0.8064   \\
 \textbf{0.270}   &\textbf{0.8093}  &   0.500   &   0.8031   \\
        0.300   &0.8073   &   0.550   &   0.8075   \\
        0.330& 0.8055     &   0.600   &   0.7954   \\
        0.370& 0.8049     &   0.650   &   0.8045   \\
        0.400& 0.8078     &   0.700   &   0.8015   \\
        0.430& 0.8042     &   0.750   &   0.8051   \\
        0.470& 0.8019      &   0.800   &   0.8027   \\
        0.500& 0.8066     &   0.850   &   0.8072   \\
        0.530& 0.8044     &   0.900   &   0.8058   \\
        0.570& 0.8028     &     &     \\
        0.600& 0.8005     &    &     \\
        0.650& 0.7911     &      &      \\
        0.700& 0.8074     &      &      \\
        0.750& 0.8018     &      &      \\
        0.800& 0.8082     &     &      \\
        0.850& 0.8022     &     &      \\
        0.900& 0.8006     &     &      \\
        0.950& 0.8076     &     &      \\
        1.000& 0.8094     &     &      \\
		\hline																
	\end{tabular}}}
\subtable[]{
\scalebox{0.7}[0.7]{
	\begin{tabular}{cccc}
		\hline
	 $\lambda$  &   accuracy  &   $\alpha$  &    accuracy \\		
		\hline
		0.001   &   0.9086    &   0.001    &  0.9093 \\
		0.005   &   0.9102    &   0.005   &  0.9087\\
		0.008   &   0.9109    &   0.010   &  0.9075 \\
		0.011   &   0.9108    &   0.050   &   0.9066\\
		0.015   &   0.9088    &   \textbf{0.100}   &   \textbf{0.9123}   \\
        0.030   &   0.9111    &   0.200  & 0.9100   \\
        0.050   &   0.9078    &   0.250   &   0.9088   \\
 \textbf{0.070}   & \textbf{0.9123}      &   0.300   &   0.9073  \\
        0.100     &0.9099       &      &    \\
        0.150    &0.9111      &      &     \\
        0.200     &0.9057      &     &     \\
        0.250    &0.9043      &      &    \\
        0.300     & 0.9035     &       &  \\
        0.350    &0.9078      &          &   \\
        0.400     &0.9061      &    &    \\
        0.450    & 0.9091     &    &    \\
        0.500     & 0.9095     &    &    \\
        0.550    & 0.9084     &    &    \\
        0.600     & 0.9070     &    &    \\
		\hline																
	\end{tabular}}
}
\end{table}

\begin{table}[H]
\centering 
\caption{The recognition accuracy of IE loss on CIFAR100 with the CIFAR100 network depicted in Tab.1, in regard to different value of $\lambda$ and $\alpha$ respectively.}\label{tableb3}
\scalebox{0.7}[0.7]{
\begin{tabular}{ lllllllllll }
\\
\hline
$\lambda$ &0.001  &0.003  &0.005  &\textbf{0.007}  &0.010 &0.030  &0.050   &0.070  &0.100   &0.200  \\
accuracy    &70.41  &71.41  &71.14 &\textbf{71.58} &70.72  &70.72  &70.85 &71.11  &70.80  &70.62\\
$\alpha$     & 0.007  &0.005   &0.001   &0.010  &\textbf{0.100}  &0.200&&&&\\
accuracy    &71.06   &70.97   &71.16   &70.82 &\textbf{71.58} &71.18&&&&\\
\hline
\end{tabular}
}
\end{table}

Here we describe the accuracy results about different hyperparameters and the optimal settings on object recognition using the little and normal networks in details. All the experiments in this part obey the following steps. First, we fix $\alpha$ to 0.1 and vary $\lambda$ according to its corresponding range in different databases. Then, we fix $\lambda$ to the best setting from the previous results and vary $\alpha$ to find the final optimal setting. Both the optimal values of $\lambda$ and $\alpha$ are displayed in bold.

\section*{References}

\bibliographystyle{ieeetr}
\bibliography{mybibfile}

\end{document}